\documentclass[11pt,a4paper]{article}
\usepackage[hyperref]{emnlp-ijcnlp-2019}
\usepackage{times}
\usepackage{latexsym}
\usepackage{booktabs}
\usepackage{amsmath}
\usepackage{amsfonts}
\usepackage{amssymb}
\usepackage{bbm}
\usepackage{mathtools}
\usepackage{graphicx}
\usepackage{array}
\usepackage{enumitem}
\usepackage{multirow}

\newcolumntype{?}{!{\vrule width 1.5pt}}
\newcommand{\hh}{\boldsymbol{h}}
\newcommand{\pp}{\boldsymbol{p}}
\newcommand{\bb}{\boldsymbol{b}}
\newcommand{\uu}{\boldsymbol{u}}
\newcommand{\vv}{\boldsymbol{v}}
\newcommand{\ff}{\boldsymbol{f}}
\newcommand{\tp}[1]{\vspace{0.1cm}\noindent\textbf{#1}}
\newcommand{\tpf}[1]{\noindent\textbf{#1}}

\newcommand{\specialcell}[2][c]{%
  \begin{tabular}[#1]{@{}l@{}}#2\end{tabular}}

\usepackage{url}

\aclfinalcopy 

\newcommand{\textff}[1]{\texttt{#1}}

\title{Working Hard or Hardly Working:\\Challenges of Integrating Typology into Neural Dependency Parsers}

\author{Adam Fisch$^*$ \quad Jiang Guo$^*$ \quad Regina Barzilay \\
  Computer Science and Artificial Intelligence Laboratory\\
  Massachusetts Institute of Technology \\
  {\tt \{fisch,jiang\_guo,regina\}@csail.mit.edu} \\}

\date{}

\begin{document}
\maketitle
\renewcommand{\thefootnote}{\fnsymbol{footnote}}
\footnotetext[1]{The first two authors contributed equally.}
\renewcommand{\thefootnote}{\arabic{footnote}}

\begin{abstract}
This paper explores the task of leveraging typology in the context of cross-lingual dependency parsing. While this linguistic information has shown great promise in pre-neural parsing, results for neural architectures have been mixed. The aim of our investigation is to better understand this state-of-the-art. Our main findings are as follows: 1) The benefit of typological information is derived from coarsely grouping languages into syntactically-homogeneous clusters rather than from learning to leverage variations along individual typological dimensions in a compositional manner; 2) Typology consistent with the actual corpus statistics yields better transfer performance; 3) Typological similarity is only a rough proxy of cross-lingual transferability with respect to parsing.\footnote{Code: \url{github.com/ajfisch/TypologyParser}}
\end{abstract}

\section{Introduction}

 Over the last decade, dependency parsers for resource-rich languages have steadily continued to improve. In parallel, significant research efforts have been dedicated towards advancing cross-lingual parsing. This direction seeks to capitalize on existing annotations in resource-rich languages by transferring them to the rest of the world's over 7,000 languages~\cite{bender2011}. The NLP community has devoted substantial resources towards this goal, such as the creation of universal annotation schemas, and the expansion of existing treebanks to diverse language families. Nevertheless, cross-lingual transfer gains remain modest when put in perspective: 
 the performance of 
 transfer models can often be exceeded using only a handful of annotated sentences in the target language (Section~\ref{results}). The considerable divergence of language structures proves challenging for current models.

One promising direction for handling these divergences is linguistic typology.
Linguistic typology classifies languages according to their structural and functional features. 
 By explicitly highlighting specific similarities and differences in languages' syntactic structures, typology holds great potential for facilitating cross-lingual transfer~\cite{ohoran2016}.  Indeed, non-neural parsing approaches have already demonstrated empirical benefits of typology-aware models~\cite{naseem2012selective, tackstrom-etal-2013-target, zhang2015hierarchical} 
 While adding discrete typological attributes is straightforward for traditional feature-based approaches, for modern neural parsers finding an effective implementation choice is more of an open question. Not surprisingly, the reported results have been mixed. For instance, \newcite{ammar2016many} found no benefit to using typology for parsing when using a neural-based model, while \newcite{wang2018surface} and \newcite{scholivet2019typological} did in several cases.

There are many possible hypotheses that can attempt to explain the state-of-the-art. Might neural models already implicitly learn typological information on their own? Is the hand-specified typology information sufficiently accurate --- or provided in the right granularity --- to always be useful? How do cross-lingual parsers use, or ignore, typology when making predictions? Without understanding answers to these questions, it is difficult to develop a principled way for robustly incorporating linguistic knowledge as an inductive bias for cross-lingual transfer.

In this paper, we explore these questions in the context of two predominantly-used typology-based neural architectures for delexicalized dependency parsing.\footnote{We focus on delexicalized parsing in order to isolate the effects of syntax by removing lexical influences.} The first method implements a variant of selective sharing~\cite{naseem2012selective}; the second adds typological information as an additional feature of the input sentence. Both models are built on top of the popular Biaffine Parser~\cite{dozat_deep_2017}. We study model performance across multiple forms of typological representation and resolution.

Our key findings are as follows:

\begin{itemize}[leftmargin=*]
    \vspace{-0.5em}
    \item \textbf{Typology as Quantization} Cross-lingual parsers use typology to coarsely group languages into syntactically-homogeneous 
clusters, yet fail to significantly capture finer
distinctions or typological feature compositions. Our results indicate that they primarily take advantage of the simple geometry of the typological space (e.g. language distances), rather than specific variations in individual typological dimensions (e.g. \texttt{SV} vs. \texttt{VS}).

    \vspace{-0.5em}
    \item \textbf{Typology Quality} Typology that is consistent with the actual corpus statistics results in better transfer performance, most likely by capturing a better reflection of the typological variations within that sample. Typology granularity also matters. Finer-grained, high-dimensional representations prove harder to use robustly.
    
    \vspace{-0.5em}
    \item \textbf{Typology vs. Parser Transferability} Typological similarity only partially explains cross-lingual transferability with respect to parsing. The geometry of the typological space does not fully mirror that of the ``parsing'' space, and therefore requires task-specific refinement.
\end{itemize}

\section{Typology Representations}
 \tpf{Linguistic Typology, $\mathbf{T}_L$:} The standard representation of typology is sets of annotations by linguists 
 for a variety of language-level properties. These properties can be found in online databases such as The World Atlas of Language Structures (WALS) \cite{wals}.
We consider the same subset of features related to word order as used by \newcite{naseem2012selective}, represented as a $k$-hot vector $T \in \{0, 1\}^{\sum_f |V_f|}$, where $V_f$ is the set of values feature $f$ may take.

\tp{Liu Directionalities, $\mathbf{T}_D$:} \newcite{liu2010dependency} proposed using a real-valued vector $T \in [0, 1]^r$ of the average \emph{directionalities} of each of a corpus' $r$ dependency relations as a typological descriptor.
These serve as a more fine-grained alternative to linguistic typology.  Compared to WALS, there are rarely missing values, and the \emph{degree} of dominance of each dependency ordering is directly encoded --- potentially allowing for better modeling of local variance within a language. It is important to note, however, that true directionalities require a parsed corpus to be derived; thus, they are not a realistic option for cross-lingual parsing in practice.\footnote{Though \newcite{wang2017fine} indicate that they can be predicted from unparsed corpora with reasonable accuracy.} Nevertheless, we include them for completeness.

\tp{Surface Statistics, $\mathbf{T}_S$:} It is possible to derive a proxy measure of typology from part-of-speech tag sequences alone. \newcite{wang2017fine} found surface statistics to be highly predictive of language typology,
while \newcite{wang2018surface}  replaced typological features entirely with surface statistics in their augmented dependency parser. Surface statistics have the advantage of being readily available and are not restricted to narrow linguistic definitions, but are less informed by the true underlying structure. We compute the set of hand-engineered features used in \cite{wang2018surface}, yielding a real-valued vector $T \in [0, 1]^{2380}$.

\section{Parsing Architecture}

We use the graph-based Deep Biaffine Attention neural parser of \cite{dozat_deep_2017} as our baseline model.
Given a delexicalized sentence $s$ consisting of $n$ part-of-speech tags, the Biaffine Parser embeds each tag $\pp_i$, and encodes the sequence with a bi-directional LSTM to produce tag-level contextual representations $\hh_i$. Each  $\hh_i$ is then mapped into head- and child-specific representations for arc and relation prediction, $\hh_i^\text{arc-dep}$, $\hh_i^\text{arc-head}$, $\hh_i^\text{rel-dep}$, and $\hh_i^\text{rel-head}$, using four separate multi-layer perceptrons.

For decoding, arc scores are computed as:
\begin{equation}
s_{ij}^{\text{arc}} = \left(\hh_i^{\text{arc-head}}\right)^T \left(U^{\text{arc}} \hh_j^{\text{arc-dep}} + \bb^{\text{arc}}\right)
\end{equation}
while the score for dependency label $r$ for edge $(i,j)$ is computed in a similar fashion:
\begin{equation}
\begin{split}
s_{(i,j),r}^{\text{rel}} = &\left( \hh_i^{\text{rel-head}}\right)^T U_r^{\text{rel}} \hh_j^{\text{rel-dep}} \ + \\
&\left( \uu_r^{\text{rel-head}}\right)^T\hh_i^{\text{rel-head}} \ + \\
&\left( \uu_r^{\text{rel-dep}}\right)^T\hh_j^{\text{rel-dep}} + b_r
\end{split}
\end{equation}
Both $s_{ij}^{\text{arc}}$ and $s_{(i,j),r}^{\text{rel}}$ are trained greedily using cross-entropy loss with the correct head or label.
At test time the final tree is composed using the Chu-Liu-Edmonds (CLE) maximum spanning tree algorithm \cite{chu1965shortest, edmonds}.

\begin{table*}[!t]
\tiny
\centering
\vspace{-0.5em}
\resizebox{\textwidth}{!}{%
\def\arraystretch{1}%
\begin{tabular}{@{}l|cc?c|c|ccc|c@{}}
\toprule
Language   & B$^*$     & +$\mathbf{T}_S^*$       & Our Baseline        & Selective Sharing      & +$\mathbf{T}_L$    & +$\mathbf{T}_D$     & +$\mathbf{T}_S$           & Fine-tune \\ \midrule
Basque     &  49.89  &   54.34         &   56.18         &  56.54                  &  56.35\rlap{$^\dagger$}    & 56.77   &   56.50   & 60.71           \\
Croatian   &  65.03  &   67.78         &   74.86         &  75.23                  &  74.07                     & 77.39   &   75.20   & 78.39           \\
Greek      &  65.91  &   68.37         &   70.09         &  70.49                  &  68.05                     & 71.66   &   70.47   & 73.35           \\
Hebrew     &  62.58  &   66.27         &   68.85         &  68.61                  &  72.02                     & 72.75   &   69.21   & 73.88           \\
Hungarian  &  58.50  &   64.13         &   63.81         &  64.78                  &  70.28                     & 66.40   &   64.21   & 72.50           \\
Indonesian &  55.22  &   64.63         &   63.68         &  64.96                  &  69.73                     & 67.73   &   66.25   & 73.34           \\
Irish      &  58.58  &   61.51         &   61.72         &  61.49\rlap{$^\dagger$} &  65.88                     & 66.49   &   62.21   & 66.76           \\
Japanese   &  54.97  &   60.41         &   57.28         &  57.80                  &  63.83                     & 64.28   &   57.04   & 72.72           \\
Slavonic   &  68.79  &   71.13         &   75.18         &  75.17\rlap{$^\dagger$} &  74.65                     & 74.17   &   75.16\rlap{$^\dagger$}   & 73.11          \\
Persian    &  40.38  &   34.20         &   53.87         &  53.61                  &  45.14                     & 56.72   &   53.03   & 59.92           \\
Polish     &  72.15  &   76.85         &   76.01         &  75.93\rlap{$^\dagger$} &  79.51                     & 71.09   &   76.29   & 77.78           \\
Romanian   &  66.55  &   69.69         &   73.00         &  73.40   &  75.20    & 76.34   &   73.82   & 75.15           \\
Slovenian  &  72.21  &   76.06         &   81.21         &  80.99   &  81.39    & 81.36   &   80.92   & 82.43           \\
Swedish    &  72.26  &   75.32         &   79.39         &  79.64   &  80.28    & 80.10   &   79.22   & 81.29           \\
Tamil      &  51.59  &   57.53         &   57.81         &  58.85   &  59.70    & 60.37   &   58.39   & 62.94           \\
\midrule
Average    & 60.97   &   64.55         &   67.53         &  67.83   &  69.07    & 69.57   &   67.86   & 72.28           \\ \bottomrule
\end{tabular}
}
\caption{A comparison of all methods on held-out test languages. UAS results are reported over the \emph{train} splits of the held-out languages, following~\cite{wang2018surface}. B$^*$ and +$\mathbf{T}_S^*$ are the baseline and surface statistics model results, respectively, of \cite{wang2018surface}.\protect\footnotemark
~\emph{Fine-tune} is the result of adapting our baseline model using only 10 sentences from the target language. All of our reported numbers are the average of three runs with different random seeds. Results with differences that are statistically \emph{insignificant} compared to the baseline are marked with $\dagger$ (arc-level paired permutation test with $p \geq 0.05$).}
\vspace{-1em}
\label{tab:main-result}
\end{table*}

\section{Typology Augmented Parsing}

\tpf{Selective Sharing:} \newcite{naseem2012selective} introduced the idea of \emph{selective sharing} in a generative parser, where the features provided to a parser were controlled by its typology. The idea was extended to discriminative models by \newcite{tackstrom-etal-2013-target}.
For neural parsers which do not rely on manually-defined feature templates, however, there isn't an explicit way of using selective sharing. Here we choose to directly incorporate selective sharing as a bias term for arc-scoring:
\begin{equation}
s_{ij}^{\text{arc-aug}} = s_{ij}^{\text{arc}} + \vv^\top\ff_{ij}
\end{equation}
where $\vv$ is a learned weight vector and $\ff_{ij}$ is a feature vector engineered using \citeauthor{tackstrom-etal-2013-target}'s  head-modifier feature
templates (Appendix B).


\tp{Input Features:} We follow \newcite{wang2018surface} and encode the typology for language $l$ with an MLP, and concatenate it with each input:
\begin{gather}
\Phi = W_2\cdot \tanh\left(W_1 \cdot \mathbf{T}^{(l)} + \bb\right) \\
\hh = \texttt{BiLSTM}\left(\{\pp_1 \oplus \Phi, \ldots, \pp_n \oplus \Phi\}\right)
\end{gather}
This approach assumes the parser is able to learn to use information in $\mathbf{T}^{(l)} \in \{\mathbf{T}_{L}^{(l)}, \mathbf{T}_{D}^{(l)}, \mathbf{T}_{S}^{(l)}\}$ to induce some distinctive change in encoding $\hh$.

\section{Experiments}
\label{results}
 \footnotetext{\newcite{wang2018surface}'s final $\mathbf{T}_S^*$ also contains additional neural features that we omitted, as we found it to underperform using only hand-engineered features.}

\tpf{Data:} We conduct our analysis on the Universal Dependencies v1.2 dataset~\cite{udv1} and follow the same train-test partitioning of languages as \newcite{wang2018surface}. We train on 20 treebanks and evaluate cross-lingual performance on the other 15; test languages are shown in Table~\ref{tab:main-result}.\footnote{Two treebanks are excluded from evaluation, following the setting of~\citet{wang2018surface}.} We perform hyper-parameter tuning via 5-fold cross-validation on the training languages. 


\tp{Results:} Table~\ref{tab:main-result} presents our cross-lingual transfer results. Our baseline model improves over the benchmark in \cite{wang2018surface} by more than $6$\%. As expected, using typology yields mixed results. Selective sharing provides little to no benefit over the baseline. Incorporating the typology vector as an input feature is more effective, with the Liu Directionalities ($\mathbf{T}_D$) driving the most measurable improvements --- achieving statistically significant gains on $13/15$ languages. The Linguistic Typology ($\mathbf{T}_L$) gives statistically significant gains on $10/15$ languages. Nevertheless, the results are still modest. Fine-tuning on only $10$ sentences yields a \textbf{2.3$\times$ larger} average UAS increase, a noteworthy point of reference.

\section{Analysis}

\begin{figure}
    \centering
    \vspace{-0.5em}
    \includegraphics[width=60mm]{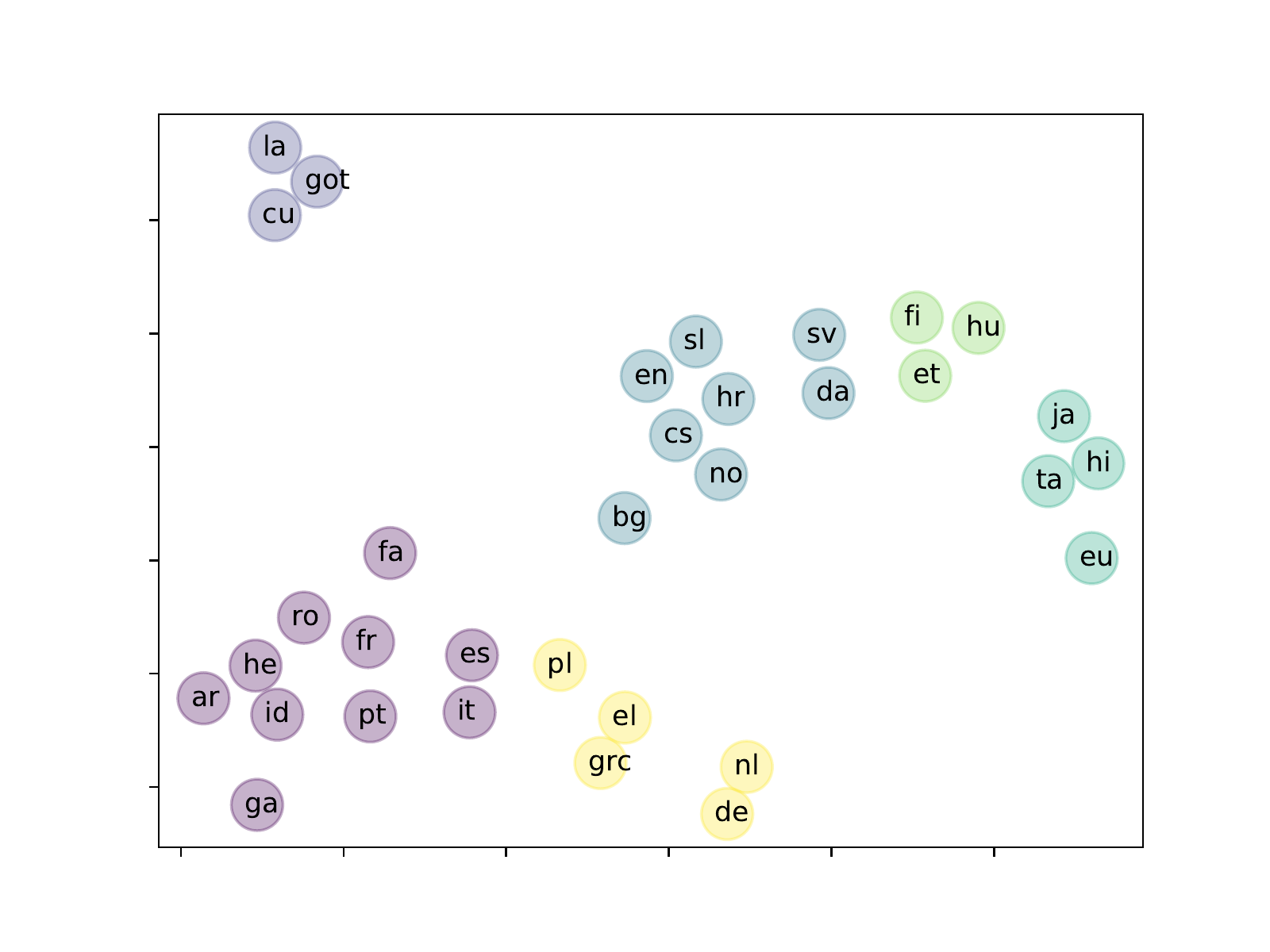}
    \caption{t-SNE projection of WALS vectors with clustering. Persian (fa) is an example of a poorly performing language that is also far from its cluster center.}
    \vspace{-0.5em}
    \label{fig:lang-distribution}
\end{figure}

\tpf{Typology as Quantization:} Adding simple, discrete language identifiers to the input has been shown to be useful in multi-task multi-lingual settings~\cite{ammar2016many, johnson2017google}. We hypothesize that the model utilizes typological information for a similar purpose by clustering languages by their parsing behavior. Testing this to the extreme, we encode languages using one-hot representations of their cluster membership. The clusters are computed by applying K-Means\footnote{We use Euclidean distance as our metric, another extreme simplification. There is no guarantee that all dimensions should be given equal weight, as indicated in Table~\ref{tab:wals-knn-acc}.} to WALS feature vectors (see Figure~\ref{fig:lang-distribution} for an illustration). 
In this sparse form, compositional aspects of cross-lingual sharing are erased. Performance using this impoverished representation, however, only suffers slightly compared to the original --- dropping by just $0.56$\% UAS overall and achieving statistically significant parity or better with $\mathbf{T}_L$ on $7/15$ languages. A gap does still partially remain; future work may investigate this further.

This phenomenon is also reflected in the performance when the original WALS features are used. Test languages that do belong to compact clusters have higher performance on average than that of those who are isolates (e.g., Persian, Basque). Indeed from Table~\ref{tab:main-result} and Fig.~\ref{fig:lang-distribution} we observe that the worst performing languages are isolated from their cluster centers. Even though their typology vectors can be viewed as compositions of training languages, the model appears to have limited generalization ability. This suggests that the model does not effectively use individual typological features.

This can likely be attributed to the training routine, which poses two inherent difficulties: 1) the parser has few examples (entire languages) to generalize from, making it hard from a learning perspective
and 2) a na\"ive encoder can already implicitly capture important typological features within its hidden state, using only the surface forms of the input. This renders the additional typology features redundant. Table~\ref{prediction} presents the results of probing the final max-pooled output of the \texttt{BiLSTM} encoder for typological features on a \emph{sentence level}. We find they are nearly linearly separable --- logistic regression achieves greater than $90$\% accuracy on average.

\newcite{wang2018surface} attempt to address the learning problem by using the synthetic Galactic Dependencies (GD) dataset~\cite{wang2016} as a form of data augmentation. GD constructs ``new'' treebanks with novel typological qualities by systematically combining the behaviors of real languages. Following their work, we add $8,820$ GD treebanks synthesized from the $20$ UD training languages, giving $8,840$ training treebanks in total. Table~\ref{gd_results} presents the results of training on this setting. While GD helps the weaker $\mathbf{T}_S^*$ substantially, the same gains are not realized for models built on top of our stronger baseline --- in fact, the baseline only narrows the gap even further by increasing by $0.92$\% UAS overall.\footnote{Sourcing a greater number of real languages may still be helpful. The synthetic GD setting is not entirely natural, and might be sensitive to hyper-parameters.}

\begin{table}[!t]
\small
\centering
\vspace{-0.5em}
\resizebox{\linewidth}{!}{%
\begin{tabular}{@{}l|clclclclclcl@{}}
\toprule
WALS ID & 82A & 83A & 85A & 86A & 87A & 88A \\ \midrule
Logreg & 87 & 85 & 97 & 92 & 94 & 92 \\
Majority & 61 & 56 & 87 & 75 & 51 & 82 \\ \bottomrule
\end{tabular}%
}
\caption{Performance of typology prediction using hidden states of the parser's encoder, compared to a majority baseline which predicts the most frequent category.}
\label{prediction}
\end{table}

\begin{table}[!t]
\small
\centering
\resizebox{\linewidth}{!}{%
\begin{tabular}{@{}l|cc?cccc@{}}
\toprule
+GD & B$^*$ & +$\mathbf{T}_S^*$ & Baseline &  +$\mathbf{T}_L^\ddagger$  & +$\mathbf{T}_D$ & +$\mathbf{T}_S$ \\ \midrule
Average & -- & 67.11 & 68.45 & 69.23 & 68.36 & 67.12\\ \bottomrule
\end{tabular}
}
\caption{Average UAS results when training with Galactic Dependencies. The Linguistic Typology ($\mathbf{T}_L^\ddagger$) here is computed directly from the corpora using the rules in Appendix E. All of our reported numbers are the average of three runs.}
\vspace{-0.5em}
\label{gd_results}
\end{table}

\tp{Typology Quality:} The notion of typology is predicated on the idea that some language features are consistent across different language samples, yet in practice this is not always the case. For instance, \textit{Arabic} is listed in WALS as \textff{SV} (82A, \textff{Subject}$^{\overset{\text{}}{\curvearrowleft}}$\textff{Verb}), yet follows a large number of \textff{Verb}$^{\overset{\text{}}{\curvearrowright}}$\textff{Subject} patterns in UD v1.2. Fig.~\ref{fig:wals-vs-ud} further demonstrates that for some languages these divergences are significant (see Appendix F for concrete examples). Given this finding, we are interested in measuring the impact this noise has on typology utilization. Empirically, $\mathbf{T}_D$, which is consistent with the corpus, performs best. Furthermore, updating our typology features for $\mathbf{T}_L$ to match the dominant ordering of the corpus yields a slight improvement of $0.21$\% UAS overall, with statistically significant gains on $7/15$ languages.

In addition to the quality of the representation, we can also analyze the impact of its resolution. In theory, a richer, high-dimensional representation of typology may capture subtle variations. In practice, however, we observe an opposite effect, where the Linguistic Typology ($\mathbf{T}_L$) and the Liu Directionalities ($\mathbf{T}_D$) outperform the surface statistics ($\mathbf{T}_S$), with $|\mathbf{T}_L| \approx |\mathbf{T}_D| \ll |\mathbf{T}_S|$. This is likely due to the limited number of languages used for training (though training on GD exhibits the same trend). This suggests that future work may consider using targeted dimensionality reduction mechanisms, optimized for performance.

\tp{Typology vs. Parser Transferability:} The implicit assumption of all the typology based methods is that the typological similarity of two languages is a good indicator of their parsing transferability. As a measure of parser transferability, for each language we select the oracle source language which results in the best transfer performance. We then compute precision@$k$ for the nearest $k$ neighbors in the typological space, i.e. whether the best source appears in the $k$ nearest neighbors. As shown in Table~\ref{tab:wals-knn-acc}, we observe that while there is some correlation between the two, they are far from perfectly aligned. $\mathbf{T}_D$ has the best alignment, which is consistent with its corresponding best parsing performance. Overall, this divergence motivates the development of approaches that better match the two distributions.


\begin{figure}[t]
    \centering
    \vspace{-0.8em}
    \includegraphics[width=70mm]{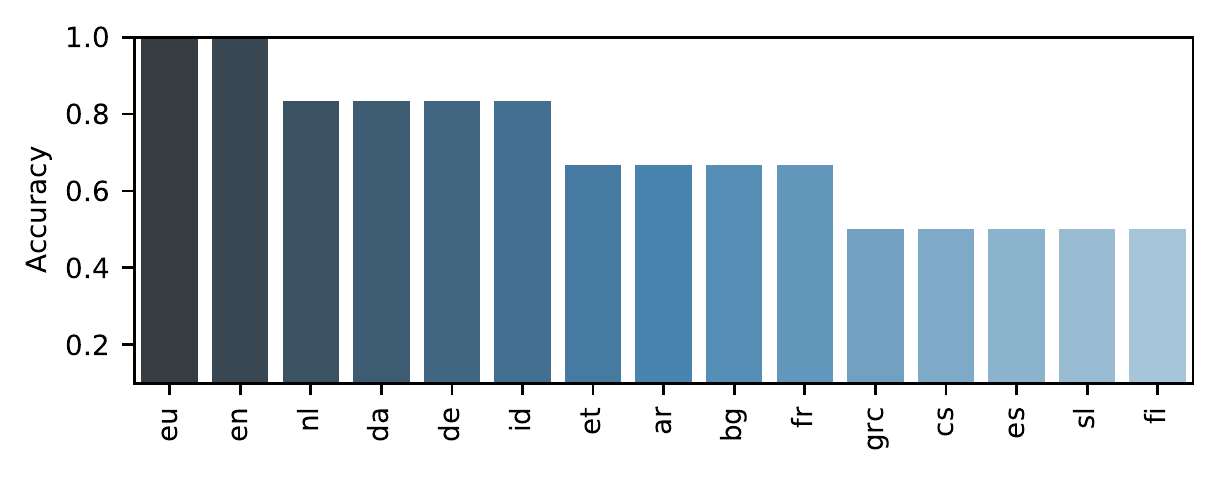}
    \vspace{-0.5em}
    \caption{Averaged matching accuracy of the linguistically-defined WALS features on 15 randomly sampled languages compared to their corpus-specific values derived from UD v1.2. Rules for deriving the features from corpus are described in Appendix E.}
    \label{fig:wals-vs-ud}
\end{figure}

\begin{table}[h]
    \small
    \centering
    \vspace{-0.5em}
    \begin{tabular}{c|cccc}
    \toprule
            & P@1 & P@3 & P@5 & P@10 \\
    \midrule
    $\mathbf{T}_L$   & 13 & 33 & 60 & 80 \\
    $\mathbf{T}_D$   & 27 & 67 & 67 & 93 \\
    $\mathbf{T}_S$   & 13 & 27 & 27 & 73 \\
    \bottomrule
    \end{tabular}
    \caption{Precision@$k$ for identifying the best \emph{parsing} transfer language, for the $k$ typological neighbors.}
    \vspace{-0.8em}
    \label{tab:wals-knn-acc}
\end{table}

\section{Related Work}

Other recent progress in cross-lingual parsing has focused on lexical alignment~\cite{guo2015cross, guo2016representation, schuster2019cross}. Data augmentation~\cite{wang2017fine} is another promising direction, but at the cost of greater training demands. Both directions do not directly address structure. \newcite{ahmad2019difficulties} showed structural-sensitivity is important for modern parsers; insensitive parsers suffer. Data transfer is an alternative solution to alleviate the typological divergences, such as annotation projection \cite{tiedemann2014rediscovering} and source treebank reordering \cite{rasooli2019low}. These approaches are typically limited by parallel data and imperfect alignments. Our work aims to understand cross-lingual parsing in the context of model transfer, with typology serving as language descriptors, with the goal of eventually addressing the issue of structure.

\section{Conclusion}
Realizing the potential for typology may require rethinking current approaches. We can further drive performance by refining typology-based similarities into a metric more representative of actual transfer quality. Ultimately, we would like to design models that can directly leverage typological compositionality for distant languages.

\section*{Acknowledgments}
We thank Dingquan Wang, Jason Eisner, the MIT NLP group (special thanks to Jiaming Luo), and the reviewers for their valuable comments.
This research is supported in part by the Office of the Director of National Intelligence, Intelligence Advanced Research Projects Activity, via contract \#FA8650-17-C-9116, and the National Science Foundation Graduate Research Fellowship under Grant \#1122374. Any opinion, findings, conclusions, or recommendations expressed in this material are those of the authors and do not necessarily reflect the views of the the NSF, ODNI, IARPA, or the U.S. Government.
We are also grateful for the support of MIT Quest for
Intelligence program.

\bibliography{emnlp-ijcnlp-2019}
\bibliographystyle{acl_natbib}

\pagenumbering{arabic} 
\appendix
\onecolumn
\begin{center}
{\bf \large{Appendix}}
\end{center}

\section{Dependency Relations for Deriving the Liu Directionalities}
\label{app:liu-relations}
Among all the 37 relation types defined in Universal Dependencies, we select a subset of dependency relations which appear in at least 20 languages, as listed in Table \ref{tab:liu-relations}. For relation types that are missing in a specific language, we simply put its value (directionality) as 0.5 without making any assumption to its tendency.

\begin{table}[h]
    \centering
    \renewcommand{\arraystretch}{1.2}
    \begin{tabular}{l}
    \toprule
    \it cc, conj, case, nsubj, nmod, dobj, mark, \\
    \it advcl, amod, advmod, neg, nummod, xcomp, \\
    \it ccomp, cop, acl, aux, punct, det, appos, \\
    \it iobj, dep, csubj, parataxis, mwe, name, \\
    \it nsubjpass, compound, auxpass, csubjpass, \\
    \it vocative, discourse \\
    \bottomrule
    \end{tabular}
    \caption{Subset of universal dependency relations used for deriving the Liu directionalities.}
    \label{tab:liu-relations}
\end{table}

\section{Feature Templates for Selective Sharing}
\label{app:selective-sharing}
We use the same set of selective sharing feature templates (Table~\ref{tab:select-sharing}) as \newcite{tackstrom-etal-2013-target}. 
\begin{table}[h]
    \centering
    \renewcommand{\arraystretch}{1.5}
    \begin{tabular}{l}
    \toprule
    \textff{d} $\otimes$ \textff{w.81A} $\otimes$ $\mathbbm{1}$\big[\textff{h.p=VERB} $\land$ \textff{m.p=NOUN}\big] \\
    \textff{d} $\otimes$ \textff{w.81A} $\otimes$ $\mathbbm{1}$\big[\textff{h.p=VERB} $\land$ \textff{m.p=PRON}\big] \\
    \textff{d} $\otimes$ \textff{w.85A} $\otimes$ $\mathbbm{1}$\big[\textff{h.p=NOUN} $\land$ \textff{m.p=ADP}\big] \\
    \textff{d} $\otimes$ \textff{w.86A} $\otimes$ $\mathbbm{1}$\big[\textff{h.p=PRON} $\land$ \textff{m.p=ADP}\big] \\
    \textff{d} $\otimes$ \textff{w.87A} $\otimes$ $\mathbbm{1}$\big[\textff{h.p=NOUN} $\land$ \textff{m.p=ADJ}\big] \\
    \bottomrule
    \end{tabular}
    \caption{Arc-factored feature templates for selective sharing. Arc direction: \textff{d} $\in$ \{\textsc{Left}, \textsc{Right}\}; Part-of-speech tag of head / modifier: \textff{h.p} / \textff{m.p}. WALS features: \textff{w.X} for \textff{X=81A} (order of \textff{Subject, Verb}  and \textff{Object}), \textff{85A} (order of \textff{Adposition} and \textff{Noun}), \textff{86A} (order of \textff{Genitive} and \textff{Noun}), \textff{87A} (order of \textff{Adjective} and \textff{Noun}).}
    \label{tab:select-sharing}
\end{table}

\section{Training details}
To train our baseline parser and its typology-augmented variants, we use \textsc{Adam}~\cite{kingma2014adam} with a learning rate of 1e-3 for 200K updates (2M when using GD). We use a batch size of 500 tokens. Early stopping is also employed, based on the validation set in the training languages. Following~\citet{dozat_deep_2017}, we use a 3-layered bidirectional LSTM with a hidden size of 400. The hidden sizes of the MLPs for predicting arcs and dependency relations are 500 and 100, respectively. 

Our baseline model shares all parameters across languages. During training, we truncate each training treebank to a maximum of 500K tokens for efficiency. Batch updates are composed of examples derived from a single language, and are sampled uniformly, such that the number of per-language updates are proportional to the size of each language's treebank. Following \cite{wang2018surface}, when training on GD, we sample a batch from a real language with probability $0.2$, and a batch of GD data otherwise. 

For  \emph{fine-tune}, we perform 100 SGD updates with no early-stopping.
When using K-Means to obtain language clusters, we set $K=5$, based on cross-validation.

\section{LAS Results}
Table \ref{tab:main-result-las} summarizes the LAS scores corresponding to Table \ref{tab:main-result} in the paper.

\begin{table*}[h]
\tiny
\centering
\resizebox{\textwidth}{!}{%
\def\arraystretch{1}%
\begin{tabular}{@{}l|cc?c|c|ccc|c@{}}
\toprule
Language   & B$^*$     & +$\mathbf{T}_S^*$       & Our Baseline        & Selective Sharing      & +$\mathbf{T}_L$    & +$\mathbf{T}_D$     & +$\mathbf{T}_S$           & Fine-tune \\ \midrule
Basque     & 27.07        & 31.46              & 34.64       & 34.79                & 36.49   & 36.83  & 34.90              & 43.04        \\
Croatian   & 48.68        & 52.29              & 61.34       & 61.41\rlap{$^\dagger$}                & 59.86   & 63.72  & 61.60              & 65.07        \\
Greek      & 50.10         & 56.73              & 56.51       & 56.53\rlap{$^\dagger$}                & 55.16   & 60.18  & 56.59\rlap{$^\dagger$}             & 64.66        \\
Hebrew     & 49.71        & 53.29              & 41.15       & 41.05                & 43.58   & 43.63  & 41.50              & 43.14        \\
Hungarian  & 42.85        & 47.73              & 32.65       & 33.43                & 34.14   & 32.01  & 33.07             & 44.26        \\
Indonesian & 39.46        & 47.63              & 47.17       & 48.21                & 51.82   & 50.78  & 49.22             & 62.23        \\
Irish      & 39.06        & 40.75              & 39.63       & 39.60\rlap{$^\dagger$}                 & 43.02   & 42.14  & 40.24             & 48.58        \\
Japanese   & 37.57        & 40.6               & 43.32       & 43.69                & 47.67   & 48.10   & 42.85             & 60.59        \\
Slavonic   & 40.03        & 43.95              & 57.35       & 57.40\rlap{$^\dagger$}                 & 56.89   & 56.69  & 57.19             & 53.88        \\
Persian    & 30.06        & 24.6               & 35.72       & 35.59                & 32.85   & 39.78  & 34.93             & 49.72        \\
Polish     & 50.08        & 54.85              & 61.67       & 61.57                & 64.69   & 57.20   & 61.71             & 65.68        \\
Romanian   & 50.90         & 53.42              & 55.77       & 56.21                & 55.99\rlap{$^\dagger$}   & 59.28  & 56.48             & 59.12        \\
Slovenian  & 57.09        & 61.48              & 70.86       & 70.01                & 70.44   & 70.03  & 70.29             & 73.81        \\
Swedish    & 55.35        & 58.42              & 67.24       & 67.40                 & 66.92   & 68.03  & 67.04             & 68.65        \\
Tamil      & 28.39        & 37.81              & 33.81       & 34.57                & 34.96   & 36.61  & 34.70              & 47.46        \\
\midrule
AVG        & 43.09        & 47.00                 & 49.26       & 49.43                & 50.30    & 51.00     & 49.49             & 56.66       
\\ \bottomrule
\end{tabular}
}
\caption{LAS results corresponding to Table 1 in the paper. Results with differences that are statistically \emph{insignificant} compared to the baseline are marked with $\dagger$ (arc-level paired permutation test with $p \geq 0.05$).}
\label{tab:main-result-las}
\end{table*}

\section{Rules for Deriving Corpus-specific WALS Features}
Table~\ref{tab:ud-wals} summarizes the rules we used to derive corpus-specific WALS features.
The values are determined by the dominance of directionalities, e.g., if $\frac{\#\{\curvearrowright\}}{\#\{\curvearrowright\}+\#\{\curvearrowleft\}} > \delta$, then its typological feature is set to the right-direction value, vice versa. In-between values are set to \textff{Mixed}. In our experiments, $\delta=0.75$.
\begin{table*}[h]
    \centering
    \small
    \renewcommand{\arraystretch}{1.2}
    \begin{tabular}{c|c|l}
    \toprule
    WALS ID & Condition & Values \\
    \midrule
    82A & \specialcell{~~~~\textff{relation} $\in$ \{\textff{nsubj}, \textff{csubj}\} $\land$\\ \textff{h.p=VERB} $\land$ (\textff{m.p=NOUN} $\lor$ \textff{m.p=PRON})} & \textff{VS($\curvearrowright$), SV($\curvearrowleft$), Mixed} \\
    \hline
    83A & \specialcell{~~~~\textff{relation} $\in$ \{\textff{dobj}, \textff{iobj}\} $\land$\\ \textff{h.p=VERB} $\land$ (\textff{m.p=NOUN} $\lor$ \textff{m.p=PRON})} & \textff{VO($\curvearrowright$), OV($\curvearrowleft$), Mixed}\\
    \hline
    85A & (\textff{h.p=NOUN} $\lor$ \textff{h.p=PRON}) $\land$ \textff{m.p=ADP} & \specialcell{\textff{Prepositions($\curvearrowleft$),} \\\textff{Postpositions($\curvearrowright$)}} \\
    \hline
    86A & \textff{h.p=NOUN} $\land$ \textff{m.p=NOUN} & \specialcell{\textff{Noun-Genitive($\curvearrowright$),}\\ \textff{Genitive-Noun($\curvearrowleft$),}\\\textff{Mixed}} \\
    \hline
    87A & \textff{h.p=NOUN} $\land$ \textff{m.p=ADJ} & \specialcell{\textff{Adjective-Noun($\curvearrowleft$),}\\ \textff{Noun-Adjective($\curvearrowright$),}\\ \textff{Mixed}} \\
    \hline
    88A & \textff{relation} $\in$ \{\textff{det}\} $\land$ \textff{m.p=DET} & \specialcell{\textff{Demonstrative-Noun($\curvearrowleft$),}\\ \textff{Noun-Demonstrative($\curvearrowright$),}\\ \textff{Mixed}}\\
    \bottomrule
    \end{tabular}
    \caption{Rules for determining the dependency arc set of each specific WALS feature type. The arc direction specificed in the parenthesis of each value indicates the global directional tendency of the corresponding typological feature.}
    \label{tab:ud-wals}
\end{table*}

\section{Examples of Mismatching between WALS and Corpus Statistics}
Table~\ref{tab:wals-vs-ud-examples} shows some examples of mismatching between WALS and corpus statistics. Substantial variations exist for some typological features, and for UD v1.2 in several cases, the dominant word order specified by linguists is questionable or even reversed (cf. Arabic subject-verb order).

\begin{table}[h]
    \centering
    \begin{tabular}{l|cc|cc}
    \toprule
    \multirow{2}{*}{Language} & \multicolumn{2}{c|}{\textsc{WALS}} & \multicolumn{2}{c}{\textsc{UD}} \\
                          & ID & Value & \#\{$\curvearrowright$\} & \#\{$\curvearrowleft$\} \\
    \midrule
    Arabic & \textff{82A} & \textff{SV} ($\curvearrowleft$) & 4,875 & 2,489 \\
    Czech  & \textff{82A} & \textff{SV} ($\curvearrowleft$) & 13,925 & 32,510 \\
    Czech  & \textff{83A} & \textff{VO} ($\curvearrowright$) & 37,034 & 20,246 \\
    Spanish & \textff{83A} & \textff{VO} ($\curvearrowright$) & 10,745 & 6,119 \\
    Finnish & \textff{86A} & \textff{G-N} ($\curvearrowleft$) & 6,010 & 8,134 \\
    \bottomrule
    \end{tabular}
    \caption{Example of mismatching between WALS and arc directionalities collected from UD v1.2. \textff{G-N} is short for \textff{Genitive-Noun}.}
    \label{tab:wals-vs-ud-examples}
\end{table}

\end{document}